\documentclass[letterpaper]{article} % DO NOT CHANGE THIS
\usepackage{aaai25}  % DO NOT CHANGE THIS
\usepackage{times}  % DO NOT CHANGE THIS
\usepackage{helvet}  % DO NOT CHANGE THIS
\usepackage{courier}  % DO NOT CHANGE THIS
\usepackage[hyphens]{url}  % DO NOT CHANGE THIS
\usepackage{graphicx} % DO NOT CHANGE THIS
\usepackage{natbib}  % DO NOT CHANGE THIS AND DO NOT ADD ANY OPTIONS TO IT
\usepackage{caption} % DO NOT CHANGE THIS AND DO NOT ADD ANY OPTIONS TO IT
\usepackage{algorithm}
\usepackage{algorithmic}

\usepackage{newfloat}
\usepackage{listings}
\DeclareCaptionStyle{ruled}{labelfont=normalfont,labelsep=colon,strut=off} % DO NOT CHANGE THIS
\floatstyle{ruled}
\newfloat{listing}{tb}{lst}{}
\floatname{listing}{Listing}
\usepackage{xspace}
\newcommand{\modelname}{CtrlProt\xspace}

\title{Controllable Protein Sequence Generation with LLM Preference Optimization}
\author{
    Xiangyu Liu\textsuperscript{\rm 1},
    Yi Liu\textsuperscript{\rm 1}, 
    Silei Chen\textsuperscript{\rm 2,3}, 
    Wei Hu\textsuperscript{\rm 1,3,}\thanks{Corresponding author}
}
\affiliations{
    \textsuperscript{\rm 1} State Key Laboratory for Novel Software Technology, Nanjing University, China\\
    \textsuperscript{\rm 2} Medical School, Nanjing University, China\\
    \textsuperscript{\rm 3} National Institute of Healthcare Data Science, Nanjing University, China\\
    \{xyl, yiliu07\}.nju@gmail.com, \{chensl, whu\}@nju.edu.cn
}

\usepackage{bibentry}
\usepackage{amsmath}
\usepackage{amssymb}
\usepackage{booktabs} % 三线表
\usepackage{makecell} % 表格内换行
\usepackage{xspace}

\usepackage[dvipsnames]{xcolor}
\usepackage{tcolorbox}
\usepackage{colortbl}
\usepackage{color}

\newcommand{\answerTODO}[1][]{\textcolor{red}{\bf [TODO]}}
\newcommand{\justificationTODO}[1][]{\textcolor{red}{\bf [TODO]}}

\begin{document}

\maketitle

\begin{abstract}
Designing proteins with specific attributes offers an important solution to address biomedical challenges.
Pre-trained protein large language models (LLMs) have shown promising results on protein sequence generation.
However, to control sequence generation for specific attributes, existing work still exhibits poor functionality and structural stability. 
In this paper, we propose a novel controllable protein design method called CtrlProt.
We finetune a protein LLM with a new multi-listwise preference optimization strategy to improve generation quality and support multi-attribute controllable generation. 
Experiments demonstrate that CtrlProt can meet functionality and structural stability requirements effectively, achieving state-of-the-art performance in both single-attribute and multi-attribute protein sequence generation.
\end{abstract}


%====================%
\section{Introduction}
% background
The objective of protein design~\cite{alphafold, structure_prediction} is to create proteins with specific biochemical functions.
Designing and exploring novel proteins offers a highly promising approach to addressing challenges in various fields, e.g., drug discovery~\cite{drug_design1,drug_design2}, vaccine and enzyme design~\cite{vaccine_design,enzyme_design}.
Generating high-quality proteins that not only possess the desired functions but also exhibit structural stability has become increasingly important.
Recently, several works based on deep models have achieved notable success in protein sequence generation~\cite{protgpt2, progen2} using protein large language models (LLMs).
They leverage the similarities between protein sequences and natural language to generate biologically relevant and functional protein sequences with high accuracy.

\begin{figure}  
\centering  
\includegraphics[width=\columnwidth]{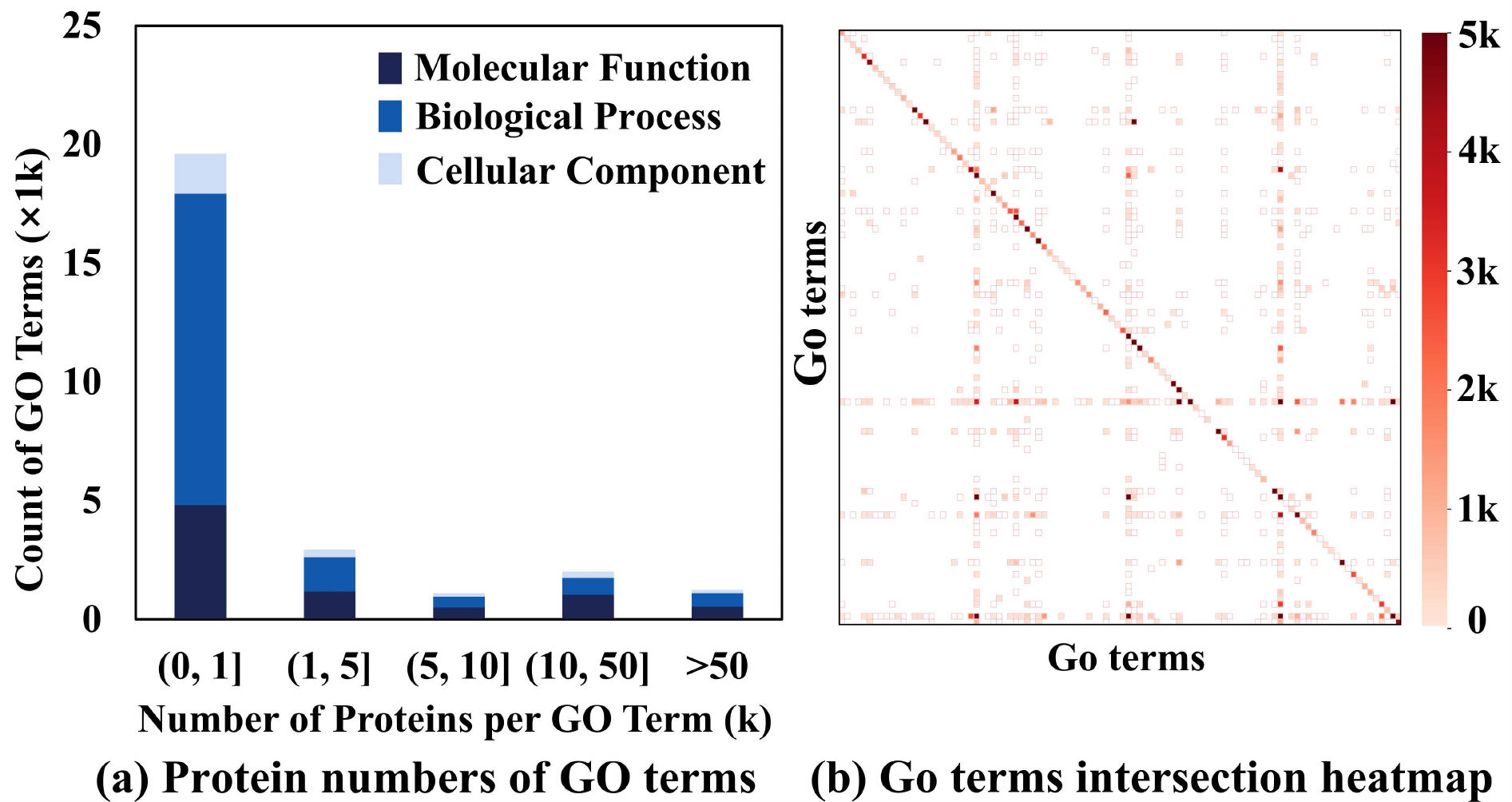}  
\caption{Gene Ontology (GO) terms analysis}  
\label{fig:data_analysis}  
\end{figure}

% existing works
Directly finetuning protein language models is an effective way to constrain their outputs to meet the specific function attributes \cite{prefixprot,tuning-plm}. 
While some works~\cite{instructprotein, moi-instructions} align protein space with natural language space to achieve controllable outputs, yielding promising results in protein sequence understanding~\cite{protllm}, finetuning on downstream tasks remains essential to ensure the generation of specific functional proteins~\cite{prollama}.

% problem & challenge
Finetuning LLMs for controllable protein generation has two major challenges. 
First, existing works based on finetuning do not explicitly constrain structural stability and functionality during training.
As a result, the finetuned model may still generate proteins that are functionally irrelevant or incapable of folding into stable structures.
Second, the effectiveness of finetuning depends heavily on the quality and quantity of the dataset.
We analyze the protein number of each term in Gene Ontology (GO) annotations\footnote{\url{https://current.geneontology.org/annotations/}}, which provides descriptions of a protein's molecular functions, biological processes, and cellular components.
As shown in Fig.~\ref{fig:data_analysis}(a), most terms have data quantities concentrated below 5,000, and there are very few terms with a sufficient number of proteins. 
Furthermore, we randomly sample 100 terms and count the number of proteins with shared attributes between any two terms.
The results in Fig.~\ref{fig:data_analysis}(b) show that fewer proteins meet both attributes, as indicated by the lighter color in the intersection of the heatmap. 
So, it is harder to construct sufficiently large data for finetuning when aiming to generate proteins with multiple attributes.

To address the issues mentioned above, we propose a preference optimization-based method to enhance the quality of controllable protein sequence generation. 
We design two metrics to assess the protein's structural stability and functionality. 
For functionality, given the critical relationship between a protein's structure and its function, we extract structural representations of sequences using a pre-trained structure encoder to assess their function similarity. 
For structural stability, we evaluate the conformational energy using the Rosetta energy score~\cite{rosetta_energy} to determine its stability.
Moreover, we propose a multi-listwise preference optimization method, which fully considers the ranking information among a large number of candidate data based on the above two metrics.
We introduce the ranking information as a regularization term to ensure that the optimization process focuses on data pairs with significant differences between preferred and non-preferred samples, thereby improving the optimization results.
Specifically, we first employ prefix-tuning on the LLM to avoid the overfitting caused by limited data. 
Then, the model generates candidate data for evaluating and constructing the preference optimization dataset.
Thus, preference optimization expands and augments the original finetuned dataset.
We further extend our method to multi-attribute generation. 
By concatenating single-attribute prefixes and applying our preference optimization, we enhance the quality of multi-attribute generation without the need of multi-attribute supervision data.

We construct the training and evaluation datasets for six attributes based on GO annotations. 
We conduct comprehensive experiments with widely-used validation metrics.
Experimental results demonstrate that our method outperforms baselines, achieving state-of-the-art results on both single-attribute and multi-attribute controllable generation.

% contribution
In summary, the main contributions of this paper are listed as follows:
\begin{itemize}
\item We propose a preference optimization-based method \modelname to enhance the quality of controllable protein sequence generation, with optimization metrics designed from functionality and structural stability perspectives.

\item We design a multi-listwise preference optimization method that leverages ranking information from large datasets, focusing on pairs with significant quality differences during training to improve the performance.

\item Experiments show that \modelname performs well in both single-attribute and multi-attribute generation tasks. 
Additionally, the detailed analyses confirm the rationality and effectiveness of \modelname.
Datasets and source code are available at \url{https://github.com/nju-websoft/CtrlProt}.
\end{itemize}

%====================%
\section{Related Work}
% In this section, we survey related work in terms of protein LLMs and preference optimization techniques.

%----------%
\subsection{Protein Language Models}
% protein language model
Protein language models are widely used in protein sequence analysis, function prediction, and protein design. 
Encoder-based protein language models like ESM~\cite{esm-1b,esm2}, ProtBERT~\cite{protbert_t5} and ProtT5~\cite{protbert_t5} are trained on large-scale protein sequence data and capable of capturing the semantic information and latent structural features within protein sequences.
Decoder-based models, such as ProtGPT2~\cite{protgpt2} and ProGen2~\cite{progen2}, have shown promising results in protein sequence generation.
Recent works also explore using diffusion models for generating protein sequences~\cite{evodiff,taxdiff}.

% controllable:1.finetune, 2.align natural language
The most straightforward method to controllable generation using protein language models is to finetune the model on downstream data~\cite{prefixprot,progen2,tuning-plm}.
It allows to generate proteins with specific functions. 
Some works aim to guide the model's generation using natural language instructions.
For example, recent studies~\cite{moi-instructions,protchatgpt,instructprotein} construct instruction-tuning datasets based on functional labels or descriptions to align models with natural language, achieving promising results in protein understanding but limited effectiveness and incomplete evaluation on sequence generation.
ProLLaMA~\cite{prollama} incorporates continual learning of protein sequences before instruction-tuning, yielding better results on generation, but it still requires finetuning on downstream data to improve performance.
However, finetuning does not explicitly focus on structural stability and functionality, which affects the quality of the generated sequences.
To address it, we propose a preference optimization method to further enhance generation quality.

%----------%
\subsection{Preference Optimization}
% rlhf ppo dpo
% listwise:optimization
The reinforcement learning from human feedback (RLHF) framework has significantly improved model performance on downstream tasks by aligning with human preferences.
Some researchers have shifted towards alternative methods to reduce the complexity of the RLHF framework.
Direct preference optimization (DPO)~\cite{dpo} directly aligns the behavior of language models without using a reward model. 
ORPO~\cite{ORPO} and SimPO~\cite{simpo} further simplify the process by eliminating the reference model.
Some works also focus on modeling sequence data. 
For example, PRO~\cite{PRO} proposes a list maximum likelihood estimation loss for response lists. 
Lipo-$\lambda$~\cite{lipo} introduces a training weight to represent the ranking difference. 
However, they are unsuitable when generating a large number of protein sequences, which is costly for PRO and may lead to an overemphasis on top sequences for Lipo-$\lambda$.
Additionally, some works~\cite{simpo,R-DPO,understanding_margin} explore the use of regularization terms as a margin to increase the reward difference between preferred and non-preferred data, thereby enhancing training effectiveness.

\begin{figure*}  
\centering  
\includegraphics[width=.9\textwidth]{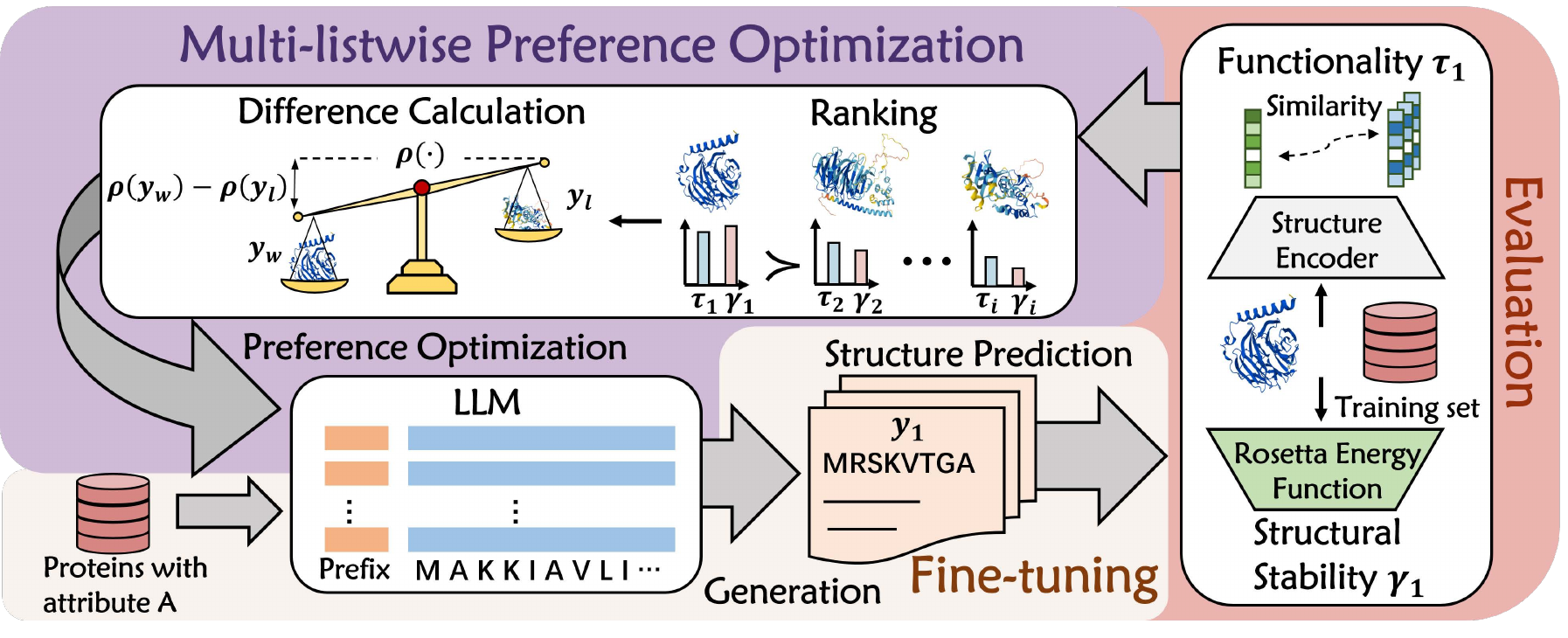}  
\caption{An overview of the proposed method \modelname.}  
\label{fig:framework}  
\end{figure*}

%====================%
\section{Preliminaries}

\paragraph{RLHF.}
In the RLHF process~\cite{RLHF,RLHF2}, an LLM is first finetuned with instructions, resulting in model $\pi_{ref}$.
The output of $\pi_{ref}$ is sampled to obtain responses $y_1,y_2 \sim \pi_{ref}$.
Humans then annotate and rank the quality of $y_1$ and $y_2$, and create the preference optimization dataset of $D = \left\{x^{(i)}, y^{(i)}_w , y^{(i)}_l \right\}^N_{i=1}$, where $y_w$ and $y_l$ denote the preferred and rejected responses, respectively.
Based on the Bradley-Terry model~\cite{BT}, the reward model $r_{\phi}(x, y)$ is used to model the preference distribution.
The training objective is
\begin{align}
\mathcal{L}(r_{\phi},D) = -\mathbb{E}_{(x, y_w, y_l) \sim D}\Big[\log \sigma \big(r_{\phi}(x, y_w) - r_{\phi}(x, y_l)\big)\Big].
\label{eq:po}
\end{align}

During the reinforcement learning phase, we treat the language model as a policy and use the trained $r_{\phi}(x,y)$ to optimize the LLM parameters. 
The optimization objective is
\begin{align}
\max\limits_{\pi_{\theta}} \mathbb{E}_{x \sim D, y \sim \pi_{\theta}(y\,|\,x)} &\big[ r_{\phi}(x, y) \big] \nonumber\\
- \beta\,\mathbb{D}_{KL} &\big[\pi_{\theta}(y\,|\,x) \,\|\, \pi_{ref}(y\,|\,x) \big].
\label{eq:rlhf}
\end{align}

\paragraph{Direct preference optimization.}
Although RLHF is effective in adapting LLMs to human preferences, it involves four sub-models, making the training complex and costly.
DPO derives a simple approach for policy optimization using preferences directly.
With the optimal solution to the KL-constrained reward maximization objective in Eq.~(\ref{eq:rlhf}), DPO rearranges it to formulate the reward function:
\begin{align}
r^*(x, y) = \beta\,\log \frac{\pi_{\theta}(y\,|\,x)}{\pi_{ref}(y\,|\,x)} + \beta\,\log Z(x),
\label{eq:reward}
\end{align}
where $Z(x) = \sum_y \pi_{ref}(y\,|\,x)\exp\big(\frac{r(x,y)}{\beta}\big)$ is the partition function. 
$r^*(x,y)$ can be substituted back into Eq.~(\ref{eq:po}), and get the following optimization objective:
\begin{align}
\mathcal{L}_{\text{DPO}}(\pi_{\theta};\pi_{\text{ref}})  = & -\mathbb{E}_{(x, y_w, y_l) \sim D} \bigg[ 
\log \sigma \bigg( 
\beta \frac{\pi_{\theta}(y_w \mid x)}{\pi_{\text{ref}}(y_w \mid x)} \nonumber \\
&\quad - \beta \frac{\pi_{\theta}(y_l \mid x)}{\pi_{\text{ref}}(y_l \mid x)} \bigg) \bigg].
\end{align}

%====================%
\section{Methods}
In this section, we provide a detailed description of our proposed method \modelname.
The framework is shown in Fig.~\ref{fig:framework}. 
The goal of \modelname is to enhance the structural stability and functionality of LLM controllable sequence generation with preference optimization. 
Our method is divided into three main steps.
First, we finetune an LLM on a protein sequence dataset with specific attributes.
Then, we generate and evaluate candidate sequences with functionality and stability metrics to build preference optimization datasets.
Finally, we perform multi-listwise preference optimization to improve the performance of the LLM.

%----------%
\subsection{Supervised Finetuning}
Suppose that we have several protein sequences related to the attribute $A$. 
We use prefix-tuning~\cite{prefix-tuning} for supervised finetuning on the attribute $A$.
Let $y = (a_1,\dots,a_l)$ be a protein sequence with $l$ amino acids.
The generation probability of $y$ can be formulated as $p(x) = \prod_{i=1}^l p(a_i\,|\,a_{<i})$ and we optimize the LLM by minimizing the negative log-likelihood as follows:
\begin{align}
L_{sft}=- \sum_{i=1}^{k}\log p_{\theta}(a_i\,|\,a_{<i},P_A),
\label{eq:sft}
\end{align}
where $P_A$ denotes the prefix related to the attribute $A$.

Although prefix-tuning can effectively leverage pre-trained knowledge and finetune data to generate sequences related to the attribute $A$, we observe from the experimental results that solely finetuning on attribute sequences may still result in protein sequences with poor structural stability and functionality. 
This indicates that finetuning alone is insufficient to fully leverage the model's understanding of protein structural semantics acquired during pre-training.
Therefore, we formulate this issue of improving the quality of the generated proteins as a preference optimization problem on structural stability and functionality.

%----------%
\subsection{DPO Data Construction}
We generate abundant candidate sequences from the prefix-tuned model $\pi_{ref}(\cdot\,|\,P_A)$. 
To fully evaluate the quality of sequence $y_i \sim \pi_{ref}(\cdot\,|\,P_A)$, we design two dimensions of evaluation metrics for stability and functionality.

\paragraph{Stability.}
The Rosetta energy function~\cite{rosetta_energy} is a widely used metric for assessing conformational energy which reflects the structural stability of the protein~\cite{progen2, protgpt2}.
It includes interactions and force fields like van der Waals forces, charge interactions, hydrogen bonds, and virtual side-chain conformations~\cite{rosetta_energy}. 
Typically, protein structures that achieve lower scores are more likely to approximate stable structures.
To compare the stability between proteins with different lengths, we normalize the raw Rosetta scores by lengths and use the per-residue Rosetta score~\cite{per_residue_rosetta} $e_i$ of $y_i$ to calculate the structural stability score $\gamma_i$:
\begin{align}
\gamma_i = 1 - \frac{e_i- e_{\min}}{e_{\max} - e_{\min}},
\label{eq:energy}
\end{align}
where $e_{\max}$ and $e_{\min}$ are the maximum and minimum Rosetta energy scores of all sequences, respectively.

\paragraph{Functionality.}
The structural similarity between protein sequences indicates their functionality relevance~\cite{structure2function}. 
We use a pre-trained encoder to obtain representations from the protein structures and measure the functionality relevance between $y_i$ and the sequences in the training set based on the similarity of their structural representations.
The functionality score $\tau_i^{A}$ is:
\begin{align}
\tau_i^{A} =\frac{1}{M}\sum_{j=1}^{M} \cos\big(\mathrm{Encoder}(y_i),\mathrm{Encoder}(y^{train}_j)\big),
\label{eq:functionality}
\end{align}
where $\cos(\cdot)$ measures cosine similarity and $y^{train}_j$ is from the training set containing $M$ sequences.
We use ProteinMPNN~\cite{proteinmpnn} as the structural encoder.

Based on the two scores $\gamma_i$ and $\tau_i^A$, we can build the preference optimization dataset $D_A$ for the attribute $A$:
\begin{align}
D_A = \left\{(y_w,y_l, \gamma_w,  \gamma_l, \tau_w^A, \tau_l^A)\,|\, \gamma_w > \gamma_l \land \tau_w^A > \tau_l^A\right\},
\label{eq:po_dataset}
\end{align}
where $y_w,y_l \sim \pi_{ref}(\cdot\,|\,P_A)$ and both scores of $y_w$ are greater than $y_l$ in $D_A$.

%----------%
\subsection{Multi-listwise DPO}
Since $D_A$ is obtained by extensive sampling and simultaneous evaluation of functionality and structural stability, our preference optimization method should also consider both multidimensionality and sequentiality.

\paragraph{Sequentiality.}
All sequences are evaluated based on determined scores, which can be compared and ranked. 
It is natural to incorporate ranking information into DPO, making the process more attentive to pairs with significant differences while reducing the intensity of optimization for pairs with minor differences~\cite{PRO}.
Here we use $G(y_i,\gamma_i)$ to calculate the weighted stability score of $y_i$:
\begin{align}
G(y_i,\gamma_i) = F(\gamma_i)(2^{\gamma_i}-1),
\label{eq:weighted_stability}
\end{align}
where $F(\cdot)$ denotes the cumulative distribution function of the distribution of $\gamma$, which provides the rank of $\gamma_i$ within the entire dataset. 
Even when scores are relatively concentrated, $F(\gamma_i)$ can still help capture and utilize the information of the subtle differences between samples.
We use the beta distribution to fit the score and calculate $F(\cdot)$.

\paragraph{Multidimensionality.}
We aim to simultaneously consider the functionality and structural stability of protein sequences during the optimization process. 
Thus, the difference between chosen and rejected sequences should be considered by $\tau^A_i$ and $\gamma_i$, respectively. 
We obtain the quality score $\rho(y_i)$ of the sequence $y_i$ by
\begin{align}
 \rho(y_i) = G(y_i,\gamma_i) + G(y_i, \tau^A_i),
\end{align}
where $G(y_i, \tau_i)$ is generated in the same way as Eq.~(\ref{eq:weighted_stability}).

\paragraph{Preference optimization.} 
We rewrite the optimization objective based on Eq.~(\ref{eq:rlhf}) with the quality score $\rho(\cdot)$:
\begin{align}
\begin{split}
\max\limits_{\pi_{\theta}} \mathbb{E}_{x \sim D, y \sim \pi_{\theta}(y\,|\,x)} &\big[ r^*(x, y) + \rho(y)\big] \\
- \beta\,\mathbb{D}_{KL} &\big[\pi_{\theta}(y\,|\,x) \,\|\, \pi_{ref}(y\,|\,x) \big],
\label{eq:rlhf_rho}
\end{split}
\end{align}
where we use $r^*(x,y) + \rho(y)$ equivalently replaces the originally hypothesized latent reward function.

Following the same derivation in DPO, we get the mapping from reward functions to optimal policies:
\begin{align}
r^*(x, y) = \beta\,\log \frac{\pi_{r^*}(y\,|\,x)}{\pi_{ref}(y\,|\,x)} + \beta\,\log Z(x) - \rho(y).
\end{align}

Finally, substituting $r^*(x,y)$ into Eq.~(\ref{eq:po}), we eliminate the partition function $Z(x)$ and obtain the final loss:
\begin{align}
&\mathcal{L}_{MLPO}(\pi_{\theta};\pi_{ref}) = - \mathbb{E}_{x, y_w, y_l \sim D}\Big[\log \sigma\Big(\beta \frac{\pi_\theta(y_w\,|\,x)}{\pi_{ref}(y_w\,|\,x)} \nonumber\\
& \qquad - \beta \frac{\pi_\theta(y_l\,|\,x)}{\pi_{ref}(y_l\,|\,x)} -\alpha\big(\rho(y_w)-\rho(y_l)\big)\Big)\Big],
\label{eq:mlpo}
\end{align}
where we add $\alpha$ to adjust the intensity.
$\alpha\,(\rho(y_w)-\rho(y_l))$ denotes the difference between preference optimization pairs and as a regularization term, influences the training. 
Specifically, when $y_w$ and $y_l$ have a significant difference, the gradient on the pairs during training will increase. 
Conversely, when $y_w$ and $y_l$ are similar in functionality and stability, the training gradient will decrease~\cite{R-DPO}. 
This can also be understood from a margin perspective~\cite{ODPO} that pairs with larger differences between $y_w$ and $y_l$ will be penalized more heavily.

\begin{table*}
\centering
{\small
\begin{tabular}{l|cccc|cccc|cccc}
\toprule & CLS & TM & RMSD & pLDDT & CLS & TM & RMSD & pLDDT & CLS & TM & RMSD & pLDDT \\ 
\midrule Models & \multicolumn{4}{c|}{MFO: Metal ion binding} & \multicolumn{4}{c|}{BPO: Phosphorylation} & \multicolumn{4}{c}{CCO: Cytoplasm}  \\
\midrule
ESM-1b & 50.1 & 54.9 & 7.17 &	68.8 &	64.5 &	60.2 &	5.29 &	68.3 &	48.4 &	52.8 &	6.74 &	57.8 \\
ESM-2 & 55.4 &	67.7 &	6.67 &	70.7 &	77.2 & \underline{75.0} &	\underline{3.31} &	70.1 &	50.4 &	62.3 &	4.27 &	60.8 \\
EvoDiff & 69.7 & 58.7 & 7.62 & 60.4 & 87.1 & 67.0 & 5.45 & 66.3 & 62.9 & 
50.5 & 7.44 & 55.6 \\
PrefixProt & 57.7 &	67.9 &	8.33 &	\underline{70.8} & 82.0 	& 67.3 	&3.77 &	\underline{76.6} &	51.1 &	47.0 	&6.77 	& 59.5 \\
ProGen2 & 62.5 & \underline{69.7} & 7.52 & 68.1 & 85.5 & 66.3 & 3.82 & 76.0 & 59.1 & \underline{63.5} & 4.45 & \underline{64.7} \\
ProLLaMA & \textbf{74.3} 	&56.9 	& \underline{5.52} 	&55.9& 	\underline{87.3} 	&66.8 &	4.79 	&64.0 	& \textbf{72.4} & 	55.9 	& \underline{3.64} 	&61.2 \\
\modelname (ours) & \underline{70.5} 	&\textbf{74.2} 	& \textbf{4.38} 	&\textbf{72.4} &	\textbf{97.4} 	&\textbf{94.1} 	&\textbf{1.18} 	&\textbf{83.8} 	& \underline{65.4} 	&\textbf{85.7} 	&\textbf{2.80} &	\textbf{81.0}  \\ 

\midrule Models & \multicolumn{4}{c|}{MFO: RNA binding} & \multicolumn{4}{c|}{BPO: Translation} & \multicolumn{4}{c}{CCO: Nucleus}  \\
\midrule
ESM-1b & 51.2 & 48.6 & 6.83 & 49.1 & 70.4 & 39.8 & 9.77 & 52.9 & 66.4 & 35.5 & 14.40 & 56.0 \\
ESM-2 & 53.9 &	38.9 &	9.19 &	52.6 &	73.3 &	46.7 &	9.12 &	51.7 &	74.8 &	\underline{37.4} &	\ \ \textbf{9.67} & 61.4 \\
EvoDiff & \underline{69.3} & 47.4 & 7.33 & 53.6 & 87.7 & 64.7 & 5.39 & 60.2 & 72.0 & 29.0 & 17.79 & 49.0 \\
PrefixProt & 67.3 &	51.7 &	9.52 &	\underline{62.7} &	83.3 	& \underline{71.0} &	5.15 &	\underline{68.7} 	&\underline{77.9} &	26.7 	&24.88 	& \underline{61.9} \\
ProGen2 & 64.2 & \underline{53.8} & \underline{6.46} & 55.6 & 87.5 & 68.5 & 6.45 & 63.8 & 76.4 & 30.2 & 16.14 & 59.8 \\
ProLLaMA & 69.2 	& 42.2 &	8.42 	& 57.9 	& \underline{88.2} 	& 59.3 	& \underline{4.82} 	& 67.2 	& 71.2 	& 23.4 	&27.50 	& 57.8 \\
\modelname (ours) &\textbf{76.1} 	&\textbf{79.3} 	&\textbf{2.86} 	&\textbf{69.9}& 	\textbf{89.0} 	&\textbf{88.2} 	&\textbf{2.54} 	&\textbf{73.0} 	&\textbf{78.4} &	\textbf{44.9} 	&\underline{12.05} 	&\textbf{63.4} \\
\bottomrule
\end{tabular}}
\caption{Results of single-attribute generation}
\label{tab:main_results}
\end{table*}

%----------%
\subsection{Generalization to Multi-attribute Generation}
Assuming we have sufficient protein sequences with multiple attributes, it can be treated as a generation task similar to the single-attribute scenario and optimized using the aforementioned method. 
However, in most cases, the amount of protein data with multi-attribute labels is limited, making it difficult to directly optimize with a multi-attribute dataset.
Therefore, we aim to extend the above method and leverage several single-attribute datasets to enhance the effectiveness of multi-attribute generation.

Suppose that we need to generate protein sequences with $K$ attributes. 
After prefix-tuning, we can get the prefix for single attribute $P_1, P_2, \dots, P_K$.
For multi-attribute generation, a naive method is to directly concatenate all the prefixes and use $P_{multi} = [P_1; P_2;\dots; P_K]$ to generate sequences with all $K$ attributes~\cite{prefixprot}.  

However, the generation quality with $P_{multi}$ is suboptimal and fails to balance multiple functionalities.
Therefore, we employ a similar preference optimization method to further refine the model.
First, we generate candidate sequences $y_1, y_2, \dots, y_n \sim \pi_{ref}(\cdot\,|\,P_{multi})$, and evaluate structural stability with Eq.~(\ref{eq:energy}) and functionality on each attribute with Eq.~(\ref{eq:functionality}).
Then, we construct the preference optimization dataset $D_{multi}$ for multi-attribute generation similar to Eq.~(\ref{eq:po_dataset}).
We can get the new quality score: 
\begin{align}
 \rho_{multi}(y_i) = G(y_i,\gamma_i) + \frac{1}{K}\sum_{k=1}^{K} G(y_i, \tau^k_i).
\label{eq:multi_weighted_score}
\end{align} 

We substitute $\rho_{multi}(y_i)$ into Eq.~(\ref{eq:mlpo}) as the final loss.

%====================%
\section{Experiments and Results}
% In this section, we first compare \modelname to state-of-the-art baselines in single-attribute generation.
% We then validate the performance of \modelname in multi-attribute generation. 

%----------%
\subsection{Experiment Setup}
\paragraph{Dataset construction.}
We extract protein sequences with Gene Ontology~(GO) terms from the UniProtKB database\footnote{\url{https://www.uniprot.org/}} and corresponding structures from the AlphaFold protein structure database\footnote{\url{https://alphafold.ebi.ac.uk/}}.
We choose six terms as attributes from three different aspects: molecular function ontology (MFO): metal ion binding and RNA binding; biological process ontology (BPO): phosphorylation and translation; cellular component ontology (CCO): cytoplasm and nucleus.
Each attribute contains 10k protein sequences for training.

\paragraph{Evaluation metrics.}
To evaluate the quality of sequences comprehensively, we use CLS-score, TM-score, and RMSD to assess their functionality and pLDDT for structural stability. 
For each attribute, we extract 100k sequences from UniProtKB as the evaluation set, excluding the training set to ensure no data leakage.
We construct a database for each attribute on the evaluation set for alignment with Foldseek~\cite{foldseek}. 
\textbf{CLS-score}: We finetune a classifier based on the ESM-2~\cite{esm2} model on the evaluation set, using the classification probabilities as the classifier score~(CLS-score). 
\textbf{TM-score} and \textbf{RMSD}: Following previous works~\cite{prollama,taxdiff}, we use Foldseek~\cite{foldseek} to assess structural similarity with Template Modeling score (TM-score)~\cite{tm-score} and Root Mean Square Distance (RMSD)~\cite{rmsd}.
Higher CLS-score or TM-score, or a lower RMSD, indicate greater structural similarity between the generated sequences and the evaluation set.
\textbf{pLDDT}: The predicted Local Distance Difference Test~(pLDDT) is used to assess the confidence of protein structure predictions. 
A higher pLDDT indicates higher prediction confidence and greater structural stability.

\paragraph{Baselines.}
We compare \modelname to six competitive baselines: 
The encoder-based methods including ESM-1b \cite{esm-1b} and ESM-2 \cite{esm2}.
The diffusion-based model is EvoDiff \cite{evodiff}.
We also compare with the state-of-the-art decoder-based methods.
PrefixProt \cite{prefixprot} uses prefix-tuning to finetune ProtGPT2 \cite{protgpt2}.
ProGen2 \cite{progen2} is pre-trained on a large corpus of protein sequences, and we finetune it using the same prefix-tuning setting.
ProLLaMa \cite{prollama} leverages the alignment of natural language to generate corresponding protein sequences.
All baselines are finetuned on the training set.

\paragraph{Settings.}
For prefix-tuning, we finetune ProtGPT2 with following settings:
batch size (16), learning rate (1e-4), prefix token number (100).
For preference optimization, we use 5k pairs on each attribute and set the learning rate (5e-5), $\beta$ (0.1), and $\alpha$ (0.05).
The maximum generation length is 400. 
We use ProteinMPNN as the structural encoder and ESMFold~\cite{esmfold} for structure prediction, both with default parameters. 
The Rosetta score is calculated using the weight configuration of ref2015~\cite{ref2015}.
All training and generation are conducted on a single A800 GPU.

%----------%
\subsection{Single-attribute Generation Results}
To assess our method \modelname's performance in controllable sequence generation, for each attribute, we generate 500 sequences using our method and baselines. 
We compare the generated results with natural proteins from the evaluation set using Foldseek.
The results, as shown in Table~\ref{tab:main_results}, indicate that our method outperforms the baselines on six single-attribute controllable generation datasets.
Notably, \modelname exhibits a significant advantage in pLDDT and TM-score, suggesting that our generated sequences have greater structural stability, and are structurally more similar to natural proteins with the same attributes, which implies similar functionality.
This verifies that \modelname effectively improves the quality of the generated proteins. 

Overall, the performance of decoder-based models after finetuning is superior to encoder-based models, highlighting their significant advantage in sequence generation tasks.
ESM-1b, ESM-2, and EvoDiff generate sequences by predicting masked tokens. It is challenging to generate protein sequences from scratch.
Therefore, we provide 10\% of amino acids to assist in the generation process.
It is worth noting that, while ProLLaMA has a larger number of parameters, its performance is not particularly remarkable.
This may be because its original instruction tuning limits the sequence length to 256. 
Although we do not impose such a restriction, it is still affected, leading to decreased performance in generating longer protein sequences.

\begin{table}
\centering
{\small
\begin{tabular}{l|cccc}
\toprule
 Variants & CLS & TM & RMSD & pLDDT \\

\midrule
 \modelname (full) & \textbf{79.5} & \textbf{77.7} & \underline{4.30} & \textbf{73.9} \\
 \ \ w/o $\gamma$ & 75.3 & \underline{75.9} & 4.41 & 70.2 \\ 
 \ \ w/o $\tau$ & 71.9 & 66.2 & 4.90 & 70.5 \\

\midrule
 \ \ w/ DPO & \underline{76.4} & 72.0 & \textbf{4.20} & 72.0 \\
 \ \ w/ ORPO & 74.0 & 70.2 & 4.80 & 72.4 \\ 
 \ \ w/ Lipo-$\lambda$ & 74.4 & 70.1 & 4.92 & \underline{73.5} \\ 
\bottomrule
\end{tabular}}
\caption{Results of ablation and alternative studies}
\label{tab:ablation_study}
\end{table}

%----------%
\subsection{Ablation and Alternative Results}
We perform ablation and alternative studies to further analyze the effectiveness of \modelname. 
As presented in Table~\ref{tab:ablation_study}, we list the average results across six single-attribute datasets. 
In the ablation study, we separately remove the functionality metric 
$\tau$ and the structural stability metric $\gamma$, and both result in a decline in \modelname's performance.
Specifically, removing $\gamma$ leads to a more significant drop in pLDDT, while removing $\tau$ causes larger decreases in CLS-score, TM-score, and RMSD.
Therefore, $\tau$ and $\gamma$ play a crucial role in enhancing functionality and structural stability, respectively.

In the alternative study, we focus on comparing several preference optimization methods: DPO, ORPO, and Lipo-$\lambda$.
Overall, our multi-listwise preference optimization outperforms all other methods.
Compared to DPO and ORPO, \modelname introduces the difference of the quality scores as a regularization term, allowing our method to pay more attention to the pairs with greater quality differences.
This leads to superior results in protein sequence data. 
On the other hand, Lipo-$\lambda$, which also employs a listwise method, performs less effectively.
This is because Lipo-$\lambda$ uses the difference of the reciprocal of ranking in its lambda weight.
When it comes to ranking and comparing in a large number of protein sequences, it overly emphasizes a few top-ranked proteins and reduces the distinction among lower-ranked proteins.
\modelname incorporates the CDF of the quality score $\rho$ to introduce ranking information and avoids this issue.

\begin{table}
\centering
{\small
\begin{tabular}{l|cc}
\toprule
    & Inter-output~(\%) & Training set~(\%)  \\
\midrule
    Natural proteins & 2.53 & 2.53 \\ 
\midrule
    ESM-1b & 6.15 & 3.13 \\
    ESM2 & 5.57 & \underline{2.60} \\
    EvoDiff & 3.39 & 2.67 \\
    PrefixProt & \textbf{3.12} & 2.73 \\
    ProGen2 & \underline{3.26} & 2.67 \\
    ProLLaMA & 5.13 & 2.73 \\
    \modelname (ours) & 3.38 & \textbf{2.55} \\
\bottomrule
\end{tabular}}
\caption{Results of diversity}
\label{tab:diversity}
\end{table}

\begin{figure} 
\centering  
\includegraphics[width=\columnwidth]{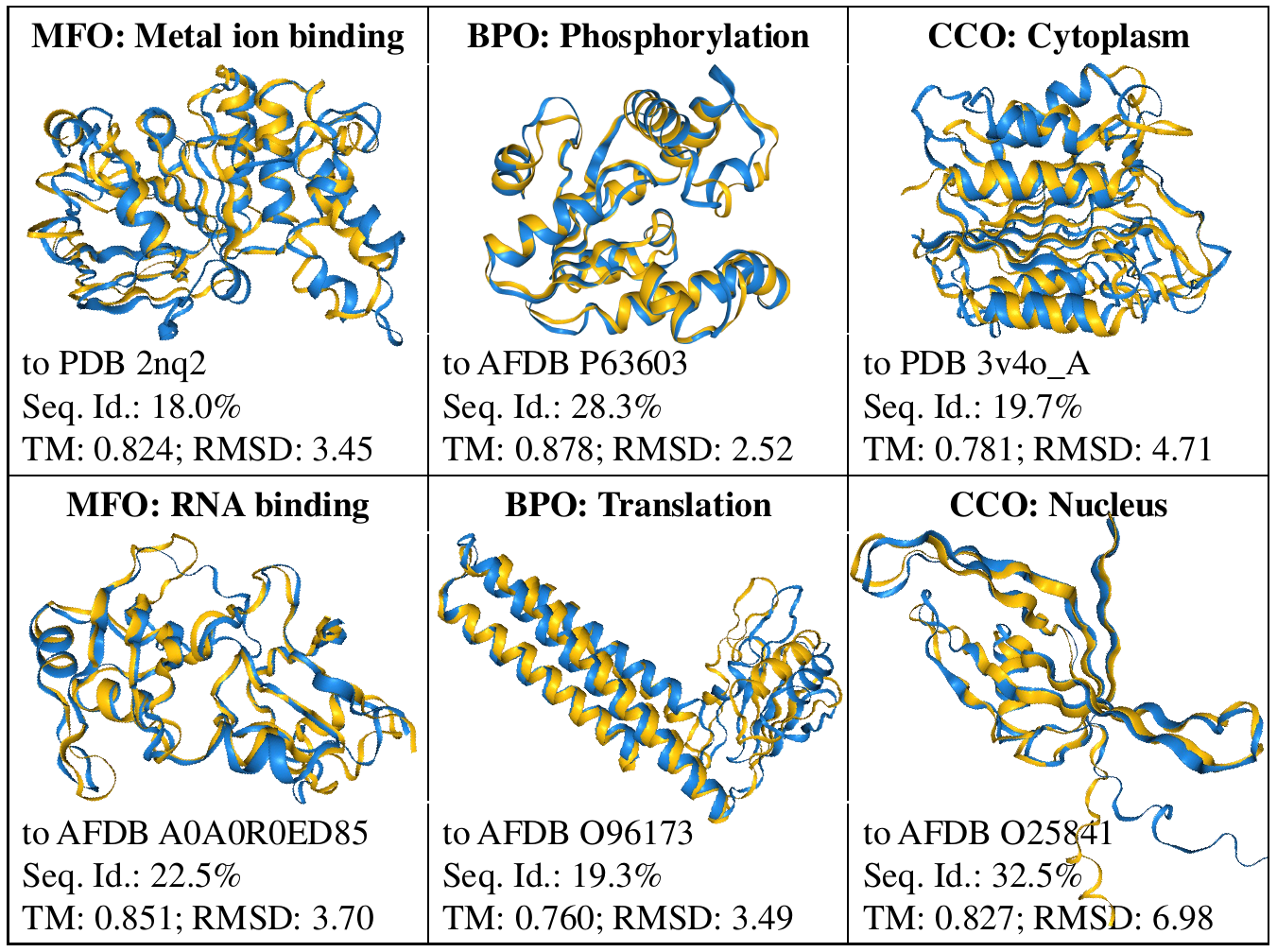}  
\caption{Case study of single-attribute generation}  
\label{fig:case_study}  
\end{figure}

\begin{figure*}[!t]  
\centering  
\includegraphics[width=.8\textwidth]{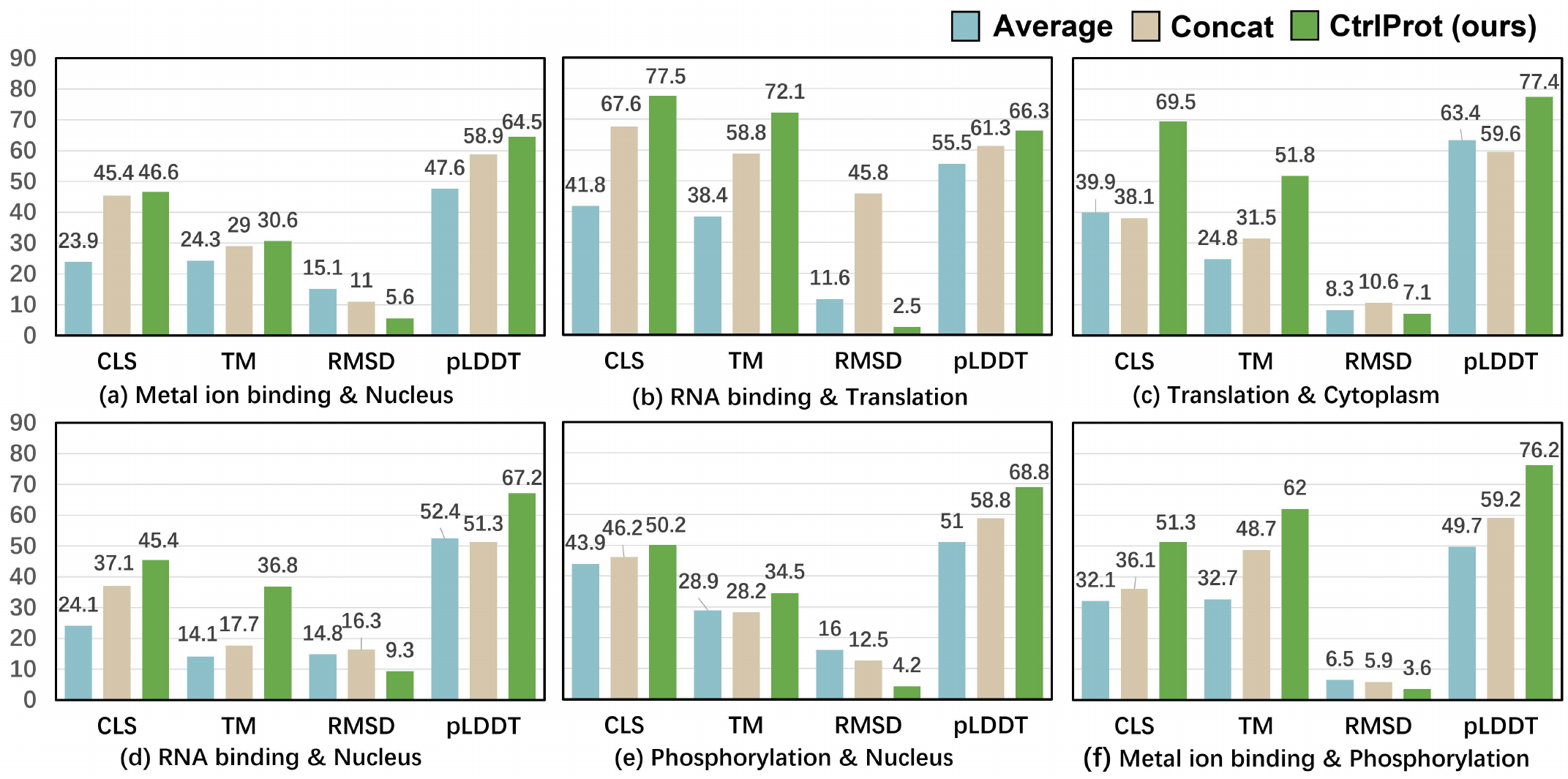}  
\caption{Results of multi-attribute generation}  
\label{fig:multi_attribute_result}  
\end{figure*}

%----------%
\subsection{Diversity Analysis}
To further verify that \modelname can generate high-quality proteins without overfitting to the training set or experiencing mode collapse~\cite{collapse}, we analyze and compare its diversity with other baselines.
Following previous works~\cite{diversity_1, diversity_2}, we use n-gram to calculate the similarity ratio between two sequences:
\begin{align}
 Sim(y_i,y_j) = \frac{|\,Set(y_i) \cap Set(y_j)\,|}{|\,Set(y_i)\,|},
\label{eq:n-gram similarity}
\end{align} 
where $Set(\cdot)$ is the set of all 3-gram items.
We report the inter-output similarity for the similarity among all generated sequences, and the training set similarity, which shows similarity between generated sequences and the training set.

The results, as shown in Table~\ref{tab:diversity}, indicate that \modelname gains the lowest similarity to the training set, close to natural proteins, which suggests that \modelname does not achieve optimal performance through overfitting.
Although the inter-output similarity of \modelname slightly increases compared to PrefixProt, it remains within a relatively optimal range.
It indicates that no mode collapse occurs during preference optimization, meaning \modelname does not resort to generating only a few patterns of proteins to obtain optimal results.

\begin{figure}  
\centering  
\includegraphics[width=\columnwidth]{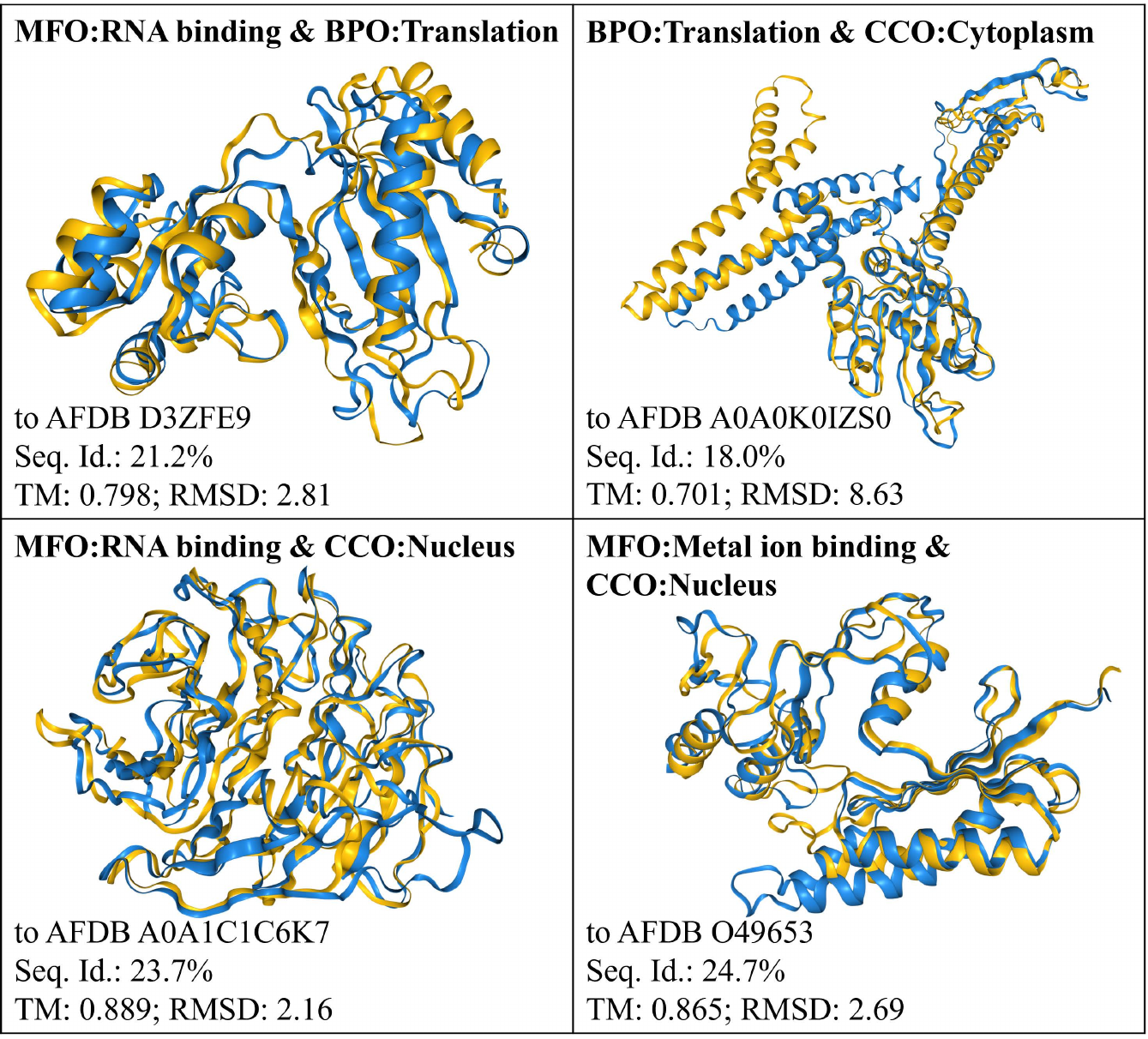}
\caption{Case study of multi attribute generation}  
\label{fig:case_study_multi}  
\end{figure}

\subsection{Case Study}
In Fig.~\ref{fig:case_study}, we present proteins generated by \modelname (shown in blue) and the most similar natural proteins (shown in yellow).
We use Sequence Identity~(Seq.Id.) from foldseek to reflect the sequence similarity.
The significant overlap in 3D structures and high TM-scores confirm structural similarity between the generated and natural proteins.
We observe that these natural proteins also satisfy the corresponding attributes, indicating functional similarity between the generated and natural proteins.
The lower Seq.Id. indicates lower amino acid sequence similarity, which means \modelname can generate desired attribute proteins with novel sequences.

%----------%
\subsection{Multi-attribute Generation Results}
\modelname can be extended to multi-attribute generation.
To validate the effectiveness, we construct six attribute combinations.
Due to the lack of studies on multi-attribute generation, we follow PrefixProt by comparing with two straightforward methods: \textbf{Average}, where the prefixes are averaged and merged, and \textbf{Concat}, where the prefixes are concatenated for generation. 
We adopt the same metrics used in single-attribute generation.
The results, as shown in Fig.~\ref{fig:multi_attribute_result}, demonstrate that \modelname achieves higher CLS-score and TM-score, as well as lower RMSD, indicating a significant improvement in functionality. 
The higher pLDDT suggests stronger structural stability in all attribute combinations.

In Fig.~\ref{fig:case_study_multi}, we present four cases of two attribute combinations.
The overlap in structures and high TM-scores shows that the generated proteins exhibit structural similarity to natural proteins with related attributes while maintaining low sequence similarity, based on low Seq.Id.

%====================%
\section{Conclusion}
In this paper, we present \modelname, a preference optimization-based method that improves the quality of controllable protein sequence generation.
We propose multi-listwise preference optimization on functionality and structural stability metrics, targeting pairs with notable quality differences.
Experiments show that \modelname excels in single-attribute and multi-attribute generation and can generate diverse proteins.
However, achieving more precise and programmable generation for certain attribute combinations remains a challenge.
We hope to continue exploring this in future work.

\section{Acknowledgments}
This work is supported by the National Natural Science Foundation of China (No. 62272219).

\bibliography{aaai25}

%%%%%%%%%%%%%%%%%%%%%%%%%%%%%%%%%%%%%%%%%%%%%%%%%%%%%%%%%%%%

\end{document}